# Working with AI

## A Design Framework for Human-AI Collaboration

**Industrial Artificial Intelligence Group**
The University of Auckland

# Contents

PREPARED BY

Industrial Artificial Intelligence Group, The University of Auckland

FUNDED BY

Transdisciplinary Ideation Fund, The University of Auckland

# Introduction

Artificial Intelligence (AI), particularly GenAI, is becoming an increasingly important part of modern work. In industrial settings, AI can support decision-making, automate routine activities, assist humans, and improve productivity. However, successful AI adoption depends on more than what the technology can do. It also depends on how people experience and work with it.

This raises an important question: how should human-AI collaboration be designed so that it works well for both people and organisations?

This white paper addresses that question by presenting a practical framework for designing human-AI collaboration. The framework considers the human, the AI system, the task, the organisation, and the wider societal environment. It explains what effective collaboration looks like, what conditions influence it, what requirements should be met, and what design decisions organisations should consider.

The report also includes a human-AI collaborative assembly system with cobot use case to demonstrate how the framework can be applied in practice. The use case shows how design requirements can be translated into specific collaboration features and evaluated through a case study.

The aim of this white paper is to provide a clear and practical guide for designing human-AI collaboration that is effective, human-centred, and responsible.

# Staff and Researchers

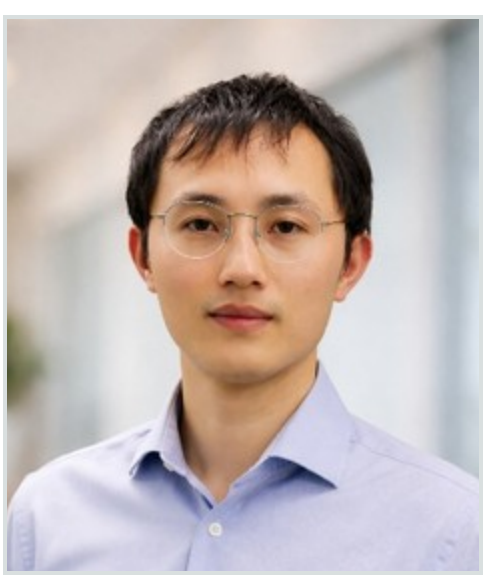

**Yuqian Lu**
**Lead Researcher**
Department of Mechanical and Mechatronics Engineering, The University of Auckland

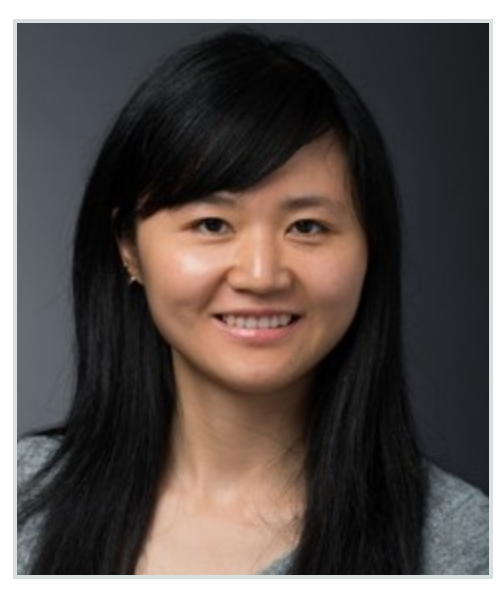

**Lixin Jiang**
**Researcher**
School of Psychology, The University of Auckland

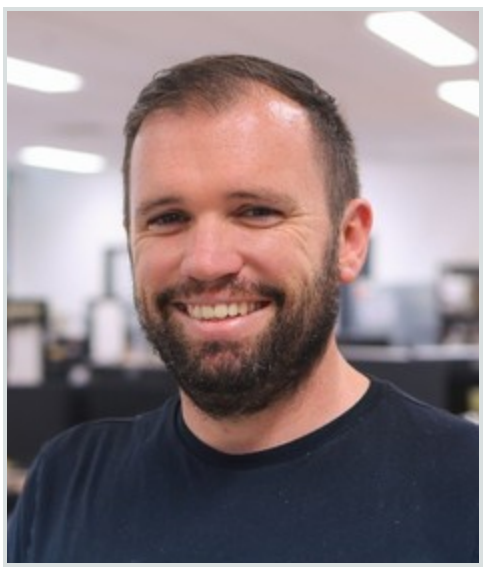

**Andrew McDaid**
**Researcher**
Department of Mechanical and Mechatronics Engineering, The University of Auckland

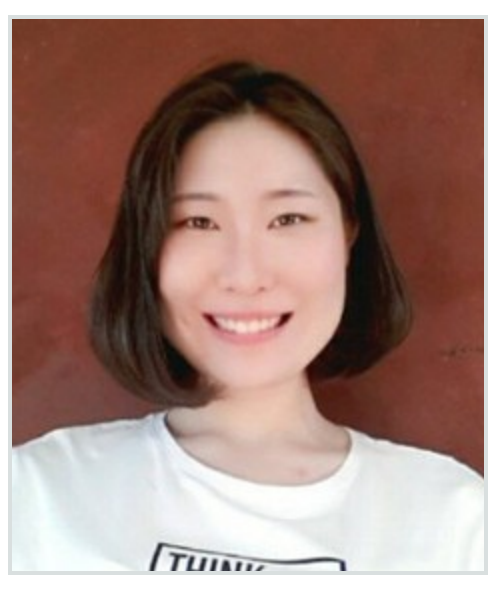

**Regina Lee**
**Graduate Researcher**
Department of Mechanical and Mechatronics Engineering, The University of Auckland

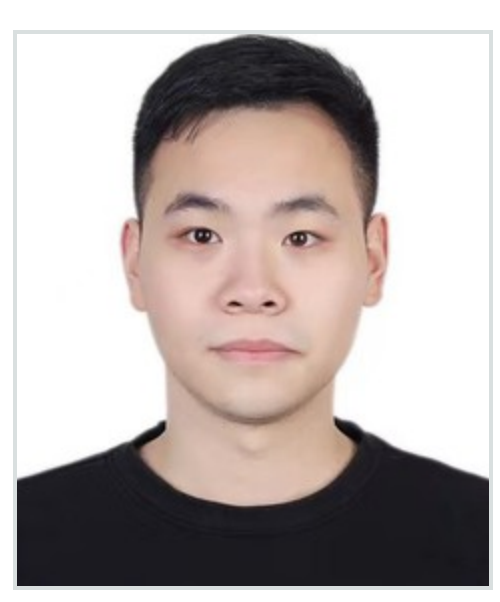

**Rui Zhou**
**Graduate Researcher**
Department of Mechanical and Mechatronics Engineering, The University of Auckland

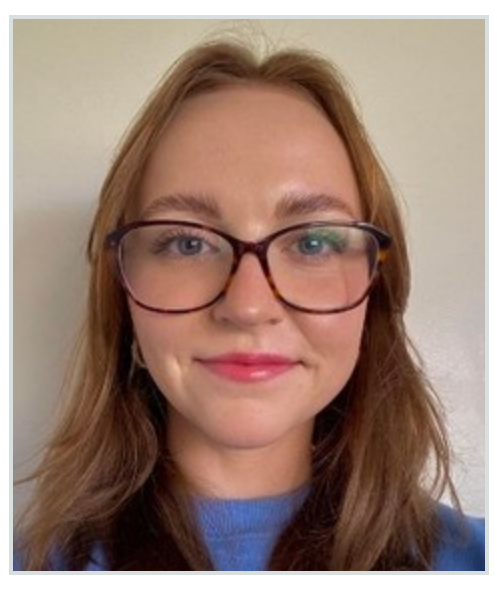

**Amy Lawrence**
**Graduate Researcher**
School of Psychology, The University of Auckland

## Supporting Partners

## How to Cite This Report

Yuqian Lu, Regina Lee, Rui Zhou, Lixin Jiang, Andrew McDaid, Amy Lawrence, “Working with AI: A Design Framework for Human-AI Collaboration,” The University of Auckland, September 2026

## Financial Acknowledgement

This report is proudly brought to life with support from the Transdisciplinary Ideation Fund at The University of Auckland.

KEY FINDINGS

# Top Takeaways

## 01 Effective human-AI collaboration must be designed, not assumed.

AI can improve productivity, creativity, and performance, but it can also create distrust, anxiety, job insecurity, role ambiguity, and loss of autonomy. Favourable outcomes depend on deliberate design, not technical capability alone.

## 02 Human-AI collaboration is a socio-technical system.

Effective collaboration depends on the interaction between the human, AI system, task, organisation, and wider societal environment. A change in one domain can affect the others.

## 03 Success is more than productivity.

Effective human-AI collaboration should support positive human experiences, strong collaboration and task performance, and sustainable organisational and societal outcomes.

## 04 Context determines the design.

Human capabilities, AI reliability, task demands, organisational support, and societal expectations shape which requirements and design decisions matter most.

## 05 Human needs and teamwork requirements are central.

Effective human-AI collaboration should support professional needs, autonomy, competence, relatedness, safety, and privacy, while enabling shared goals, coordination, communication, adaptability, calibrated trust, and appropriate human control.

## 06 Responsible deployment starts at the design stage.

Fairness, transparency, safety, privacy, security, accountability, training, and ongoing support should be built into the system from the outset rather than added after deployment.

## 07 Human-AI collaboration should be continuously evaluated and refined.

As humans gain experience and technologies or tasks change, collaboration requirements may also change. The framework therefore treats design and evaluation as an iterative process.

# 01

# Why Human-AI Collaboration Needs Deliberate Design?

CHAPTER 01

# Why Human-AI Collaboration Needs Deliberate Design?

Artificial Intelligence is increasingly used in workplaces to automate processes, support decision-making, and assist humans with complex tasks. These technologies can improve productivity and performance. However, evidence also suggests that: The same AI can create distrust, job insecurity, and cognitive demands [1], [2], [3]. Hence, AI can enhance work, but it can also introduce new human and organisational challenges.

This is because humans' responses to AI depend on more than technical capability: their responses also depend on how the AI is introduced and experienced. The same AI can be considered an assistant or a threat, depending on whether users trust it, have prior experience, and are supported through the change [1], [3], [4]. Organisational context is also important. Management support and implementation strategy also determine whether AI is accepted or resisted [3], [5]. At the technology level, the AI's reliability, transparency, controllability and safety influence how it is perceived and used [1], [3], [5].

Altogether, a technically capable system can still fail in practice without deliberate considerations of the wider context. This is why human-AI collaboration should be treated as a socio-technical design challenge [1], [3] involving the user, the AI system, the task, the organisation, and the wider societal environment [5].

This is particularly relevant to New Zealand. In 2024 Cabinet agreed a national strategic approach to AI in New Zealand [6]. The paper identified mistrust of AI, together with low uptake of AI across the economy, as key barriers to New Zealand realising the benefits of AI.

The same dynamic is at play at both levels. Trust in AI must be earned through how it is introduced and experienced, whether in a single workplace or across a national economy. Workplaces that fail to design human-AI collaboration deliberately reinforce the wider mistrust Cabinet identified as a national barrier, while that same national uncertainty makes organisations more hesitant to invest in AI in the first place. How human-AI collaboration is designed is therefore not only an organisational concern. It is a national concern.

The central premise of this white paper is therefore: **effective human-AI collaboration must be designed, not assumed**. The following sections introduce a framework for understanding what shapes collaboration, what conditions must be satisfied, what design decisions should be considered, and what outcomes should be evaluated.

# 02

# Designing Human-AI Collaboration: A Framework

IN THIS CHAPTER

CHAPTER 02

# Designing Human-AI Collaboration: A Framework

Human-AI collaboration should be understood as a socio-technical system (Figure 1). Effective collaboration depends on the relationship between five interconnected domains: the human, the AI system, the task, the organisation, and the wider societal environment. Each of these domains introduces conditions that can support or constrain collaboration, and changes in one domain can influence the others.

The framework works from two starting points as input:

- **Desired outcomes and requirements**: the human needs, human-AI teamwork, and responsible deployment conditions that need to be satisfied for collaboration to work effectively.
- **Preconditions**: the characteristics of the human, AI, task, organisation, and society that shape the collaboration.

Together, the desired outcomes and requirements, and the preconditions, inform how we should design the Human-AI Collaboration System. This is the Design Principles.

- **Design principles**: principles that guide the design decisions used to respond to the requirements to generate desired outcomes.

Once deployed, the system generates outcomes:

- **Outcomes**: the effects of human-AI collaboration on human, task and collaboration performance, organisations, and society.

Comparing the generated outcomes against the desired outcomes reveals how effective the human-AI collaboration is, and whether the system is working as intended.

The framework is intended to be used as an iterative design and evaluation process. Organisations first specify the desired outcomes and requirements arising from their use case and then identify the context and preconditions of the human-AI collaboration system. These desired outcomes and requirements guide the subsequent design decisions and serve as the objectives against which the implementation is evaluated. The results of that evaluation, in turn, inform further changes to the system, closing the loop.

This approach recognises that there is no single configuration of human-AI collaboration that will work across all workplaces. The appropriate design depends on the human, technology, task, organisational setting, and wider environment. The purpose of the framework is therefore not to prescribe one solution, but to provide an overarching framework for making context-specific and evidence-informed design decisions.

**Figure 1** **Human-AI collaboration as a socio-technical system**

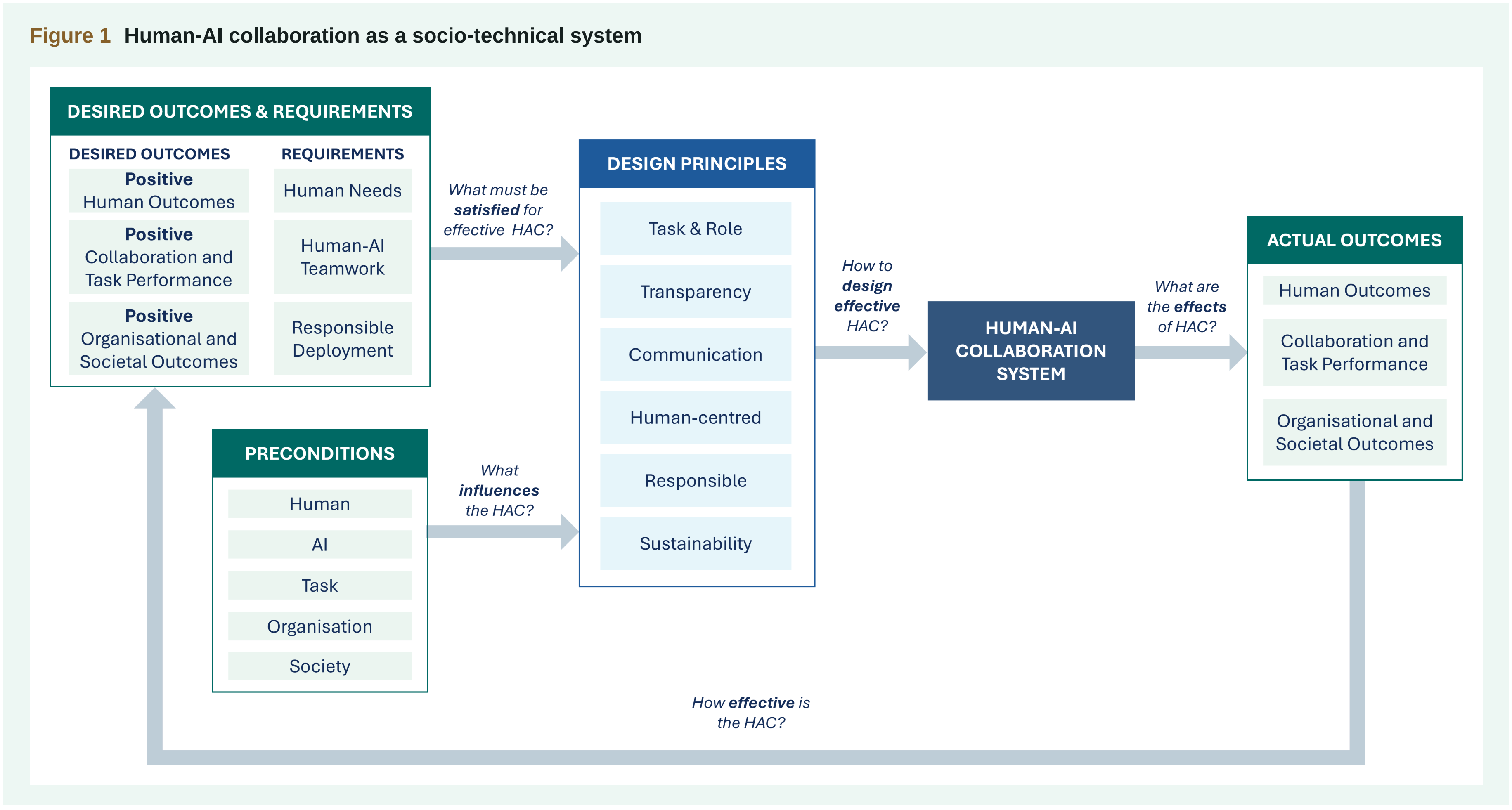

The following sections describe each part of the framework in detail, beginning with the outcomes that define effective human-AI collaboration.

# 2.1 The Outputs of Human-AI Collaboration

### What does effective human-AI collaboration look like?

Effective human-AI collaboration should produce positive outcomes at three levels:

- **Human outcomes**: how collaboration affects humans' psychological, physical, and professional experience.
- **Collaboration and task performance outcomes**: how well the human and AI work together and complete the task.
- **Organisational and societal outcomes**: the broader and longer-term consequences of the collaboration.

These three levels provide the basis for evaluating human-AI collaboration effectiveness. A system should not be considered effective simply because it improves task performance; it should also support positive human experiences and sustainable organisational and societal outcomes.

## 2.1.1 Human Outcomes

### How does the human-AI collaboration impact the human?

Human outcomes describe how human-AI collaboration affects individual psychological, professional, physical and cognitive experience. These effects can be positive or negative.

Human outcomes can be organised into four categories (Figure 2):

These categories are closely related. For example, a negative emotional response can reduce trust in the system [3], while concerns about job security or loss of autonomy can affect both acceptance and well-being [1]. Separating these categories, however, allows each to be observed and addressed in a targeted way.

Each outcome also has indicators (Table 1), which can be assessed using a combination of measurement methods. Physiological measures, such as heart rate variability and skin conductance, can provide observations for indicators such as stress and fatigue. Behavioural measures, such as task performance and interaction behaviour, can provide observations for learning curves and reliance on AI. Subjective measures, such as surveys, interviews and think-aloud protocols can provide observations for trust and sense of autonomy.

**Figure 2** Human outcomes

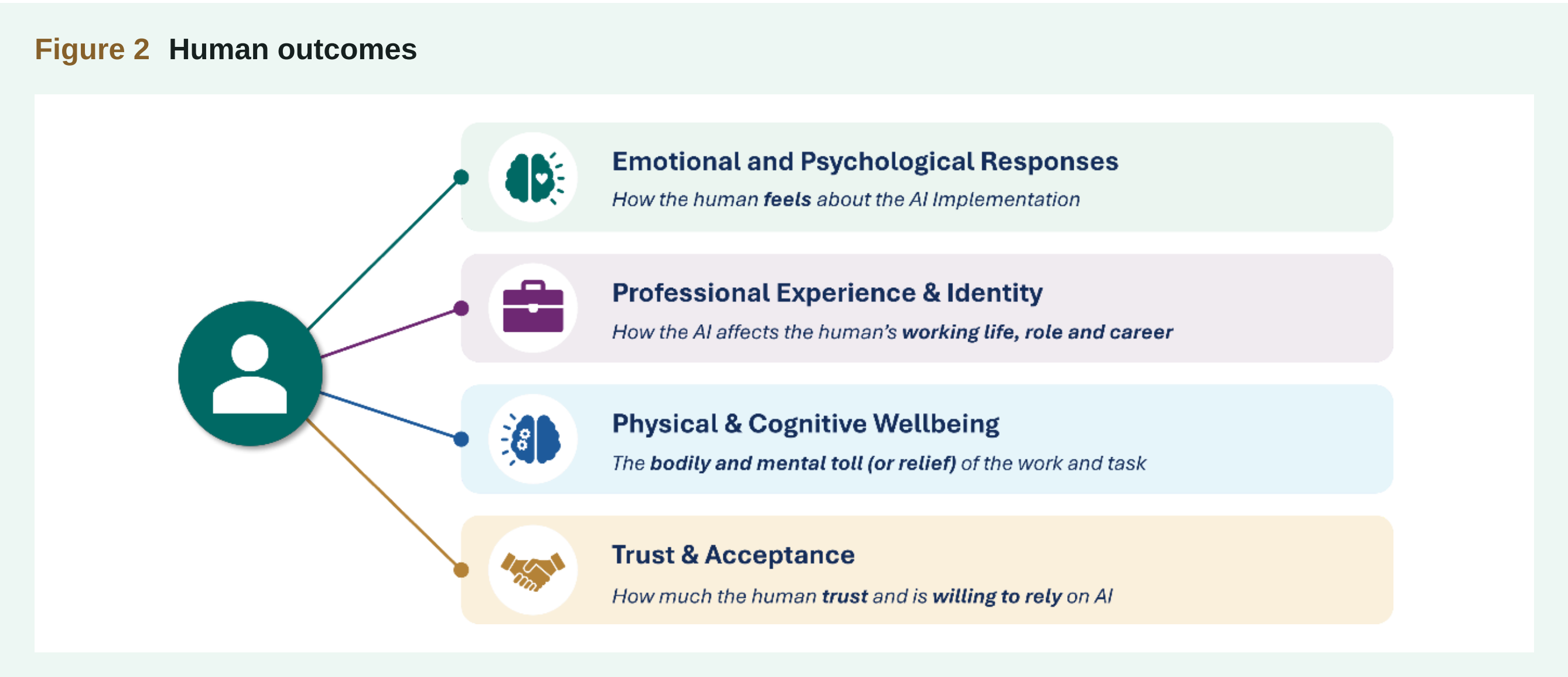

**Table 1** Human outcome indicators

| Outcome Category | Indicators | Desired outcome |
|---|---|---|
| **Emotional & Psychological Responses** | ▪ Fear [4]<br>▪ Anxiety [7]<br>▪ Frustration [8]<br>▪ Feeling of imposed cooperation [8]<br>▪ Unfamiliarity and wariness [9] | Minimise negative responses (such as fear, anxiety)<br>Maintain or improve positive responses (such as sense of control) |
| **Professional Experience & Identity** | ▪ Job (in)security [1], [3]<br>▪ Skill relevance, obsolescence [4], [10]<br>▪ Learning curve [10], [11]<br>▪ Role and task change [4], [7]<br>▪ Career development [11]<br>▪ Sense of meaning [1]<br>▪ Sense of control/autonomy [1], [2]<br>▪ Occupational status [1] | Maintain or improve professional experience and identity. |

**Table 1** Human outcome indicators (continued)

| Outcome Category | Indicators | Desired outcome |
|---|---|---|
| **Physical & Cognitive Well-being** | Physical and cognitive<br>▪ Fatigue [4], [11]<br>▪ Strain [8]<br>▪ Exhaustion [10]<br>▪ Discomfort [8]<br>▪ Stress [4], [11] | Maintain physical and cognitive well-being<br>Minimise the negative indicators. |
| **Trust & Acceptance** | ▪ Trust [3], [11]<br>▪ Reliance [9], [12]<br>▪ Satisfaction [8]<br>▪ Reliability [5]<br>▪ Accountability [1], [3]<br>▪ Constructive feedback and recognition [11] | Calibrated trust and acceptance, so that there is enough trust to work smoothly with AI, but not enough to cause overreliance. |

## 2.1.2 Collaboration and Task Performance Outcomes

### How well do the human and AI work together and perform the task successfully?

Collaboration and task performance outcomes describe how well the human and AI work together and how successfully they complete the work. They capture both the quality of the collaboration and the performance achieved through that collaboration.

These outcomes can be grouped into two categories (Figure 3).

**Figure 3** Collaboration and task performance outcomes

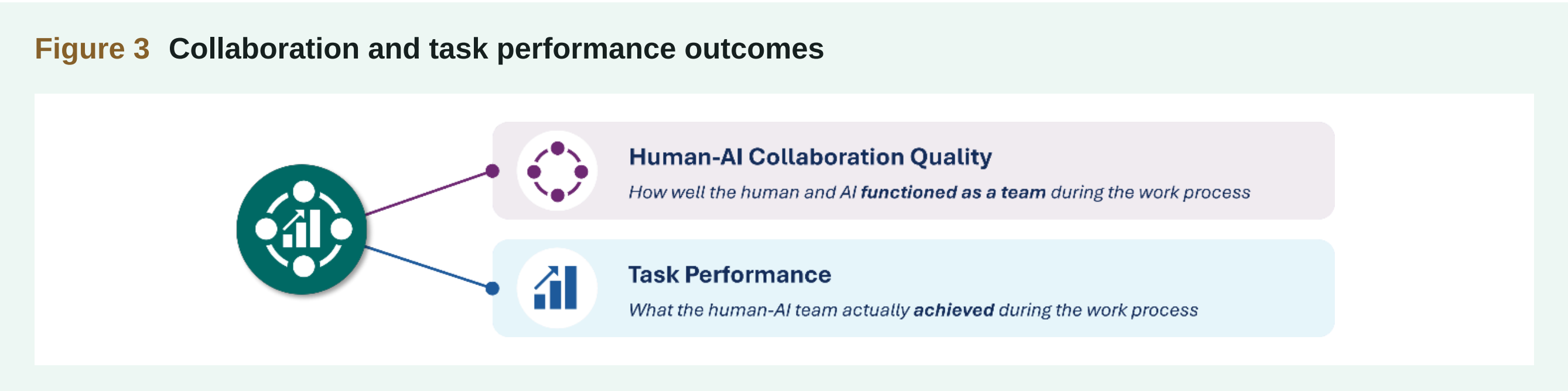

These two categories are closely connected. Strong collaboration does not necessarily guarantee strong task performance, and high task performance alone does not mean the collaboration is effective. For example, a system may complete a task quickly while creating poor coordination, excessive reliance, or an inappropriate distribution of control. Evaluating both dimensions provides a more complete picture of how well the human-AI collaboration system is functioning.

Collaboration and task performance outcomes can be assessed using both behavioural and subjective measures (Table 2). Behavioural measures such as task completion time and error rate can provide observations for task efficiency and accuracy. Other interaction behaviours such as frequency of communication and response time can provide observations for communication effectiveness and responsiveness. Subjective measures, such as surveys and think-aloud protocols, can provide observations for goal alignment and shared understanding.

**Table 2 Collaboration and task performance indicators**

| Outcome Category | Indicators | Desired outcome |
|---|---|---|
| **Human-AI Collaboration Quality** | ▪ Goal alignment [8]<br>▪ Shared understanding [13]<br>▪ Coordination [3]<br>▪ Communication effectiveness [14]<br>▪ Calibrated trust and reliance [12]<br>▪ Adaptability and responsiveness [14]<br>▪ Appropriate control [8] | Improved human-AI collaboration quality. Smooth collaboration between humans and AI. |
| **Task Performance** | ▪ Accuracy and quality [15]<br>▪ Productivity and efficiency [1]<br>▪ Consistency [14]<br>▪ Learning curve [15]<br>▪ Workload balance between human and AI [13] | Improved task performance. |

## 2.1.3 Organisational and Societal Outcomes

### What are the broader and longer-term consequences of human-AI collaboration?

Organisational and societal outcomes are the broader and longer-term effects of human-AI collaboration beyond the individuals and immediate task. These outcomes are organised into two categories (Figure 4).

**Figure 4 Organisational and societal outcomes**

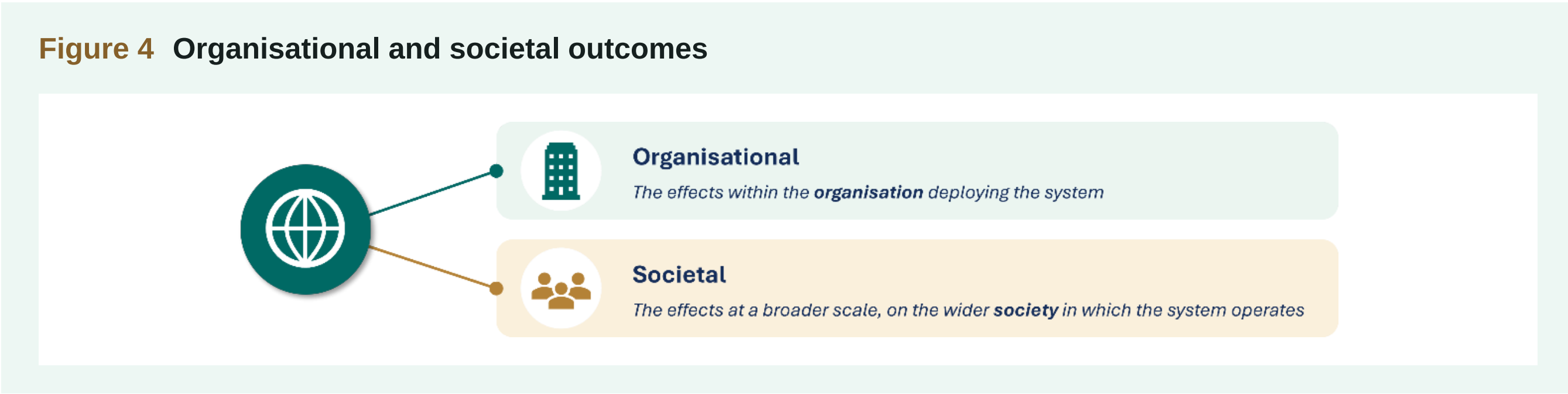

These outcomes typically emerge over time and across multiple deployments, making them important for assessing whether AI adoption is sustainable and beneficial at scale.

As a result, these outcomes are observable through organisational metrics (e.g., adoption rates, staff retention rate, ROI tracking) and societal-level research (e.g., public surveys, policy analysis, labour market data).

**Table 3** **Organisational and societal indicators**

| Outcome Category | Indicators | Desired outcome |
|---|---|---|
| **Organisational** | ▪ Productivity and efficiency gains [7]<br>▪ Return on investment [5]<br>▪ Workforce capability & retention [16], [17]<br>▪ Innovation and competitive advantage [18]<br>▪ Governance & internal accountability [1]<br>▪ Sustainable adoption [19] | Positive long-term impacts on the organisational level. |
| **Societal** | ▪ Public trust & awareness [1], [18]<br>▪ Employment & labour market effects [4]<br>▪ Privacy & surveillance concerns [11]<br>▪ Legal & regulatory accountability [3]<br>▪ Occupational standards & professional roles [4], [16]<br>▪ Prosperity [1] | Positive long-term impacts on the societal level. |

# 2.2 The Requirements for Human-AI Collaboration

### What must be satisfied to achieve the desired outcomes of human-AI collaboration?

Altogether, the human, performance and long-term outcomes describe the space of outcomes that human-AI collaboration can produce, both positive and negative. Not every outcome carries equal weight in every context. Which specific outcomes should be prioritised, in other words, which outcomes are desired, depends on the preconditions of the collaboration, discussed in the next section.

Achieving desired outcomes, however, is not automatic. It depends on a set of underlying requirements being met. The three requirement layers are:

- **Human needs**: conditions that support positive human experiences, including professional needs, psychological needs, and personal well-being.
- **Human-AI teamwork requirements**: conditions that enable the human and AI to work together effectively, such as shared goals, shared understanding, coordination, communication, adaptability, calibrated trust, and appropriate control.
- **Responsible deployment requirements**: conditions that help ensure the system remains beneficial, safe, fair, transparent, accountable, and sustainable over time.

## 2.2.1 Human needs

### What conditions must be satisfied for individual users?

Humans have needs that must be satisfied to have positive experiences. These needs can be organised into three categories.

**Table 4 Human needs**

| Needs | Description |
|---|---|
| **Professional** | Concerns the human's role and career. Examples include [1], [10]:<br>▪ Job security<br>▪ Opportunities for learning and growth<br>▪ A sense that the work remains meaningful |
| **Psychological** | Draws on the Self-Determination Theory [20]. Concerns the three basic psychological needs for wellbeing and motivation [17].<br>▪ Autonomy: a sense of control over one's actions<br>▪ Competence: feeling capable and effective<br>▪ Relatedness: a sense of connection and belonging |
| **Personal wellbeing** | Concerns the human beyond the immediate demands of the task. Examples include [11]:<br>▪ Safety<br>▪ Privacy: how their data and actions are being monitored and recorded. |

These needs are closely connected. For example, a system that improves productivity but reduces human autonomy, creates uncertainty about job security, or increases concerns about surveillance may still result in a poor human-AI collaboration experience, even when task performance outcomes look positive.

### 2.2.2 Human-AI Teamwork Requirements

#### What conditions must be satisfied for good teamwork?

To ensure smooth teamwork between the human and AI, there are functional conditions required. These teamwork requirements draw on the “Big Five” model of teamwork [21], which identifies the core components and mechanisms of effective human teams. Most of these map closely to human-AI teams. However, one difference would be in leadership/appropriate control. Compared to human teams, human-AI teams require granting the human teammates a more authoritative role over the AI for safety.

**Table 5** **Teamwork requirements**

| Needs | Description |
| --- | --- |
| **Shared goals [8]** | The human and AI should pursue the same task objective. |
| **Shared understanding [13]** | The human and AI should maintain understanding of the current task state and context, as well as each other’s intentions and capabilities. |
| **Coordination [8]** | The collaboration should consider the human to coordinate the timing, sequencing, and task allocation. |
| **Communication [14]** | Information must flow smoothly between the human and AI. This includes request, intentions, explanations, uncertainty and feedback. |
| **Adaptability [14]** | The collaboration should be able to adapt to changes in the human, task or context. It must also be resilient to failure, disagreements and unexpected events. |
| **Calibrated trust [12]** | There must be trust between the human and AI to perform the tasks as expected. However, this trust must be calibrated so that overreliance does not occur. |
| **Appropriate control [8]** | The human must be able to intervene in the AI at any time. The authority and responsibility of the human-AI team should be clear. |

These requirements are closely related. For example, effective coordination depends on communication and shared understanding, while calibrated trust depends on the human having enough information to judge when the AI should be relied on.

The design implication is that human-AI collaboration should be treated as a teamwork problem, not simply as an interface problem. The system should make roles, intentions, responsibilities, and control clear enough for the human and AI to coordinate effectively and recover when collaboration breaks down.

## 2.2.3 Responsible Deployment Requirements

### What conditions must be satisfied for human-AI collaboration deployment to be beneficial long-term?

Organisations that deploy human-AI collaboration systems must consider the conditions required to ensure the system is beneficial, responsible and sustainable in the long-term.

This can be achieved by aligning human-AI collaboration system deployment with internationally recognised principles, such as the OECD AI Principles [22]. The OECD AI Principles are designed to guide the development of trustworthy and human-centric AI. New Zealand's own 2025 Cabinet strategy on AI similarly identifies the OECD AI Principles as the country's key direction for responsible AI [6]. Because the principles are stated at a general level, the table below shows how they should be applied in responsible human-AI collaboration deployment.

**Table 6** **Responsible deployment requirements**

| OECD Principle | Description | Application to human-AI collaboration deployment |
|---|---|---|
| **Inclusive growth, sustainable development and well-being** | The potential for trustworthy AI to contribute to overall growth and prosperity for all—individuals, society, and planet—and advance global development objectives | The system should be deployed to benefit both the organisation and individuals. This includes:<br>▪ Efficiency and output gains for the organisation<br>▪ Support, training and resources for the employee's skills, working conditions and job quality |
| **Human rights and democratic values, including fairness and privacy** | AI systems should be designed in a way that respects the rule of law, human rights, democratic values and diversity, and should include appropriate safeguards to ensure a fair and just society. | The system should respect the human's autonomy and rights. This includes:<br>▪ Avoiding unfair task allocation<br>▪ Avoiding biased evaluation of performance<br>▪ Accommodate users with differing abilities, experiences and preferences<br>The system must align with regulations, public expectations and societal values. |
| **Transparency and explainability** | This principle is about transparency and responsible disclosure around AI systems to ensure that people understand when they are engaging with them and can challenge outcomes. | The system should allow the user to know when, how and why the AI is acting. Hence, with this awareness, the user should also be able to:<br>▪ Override the AI's actions<br>▪ Contribute to challenging, changing and improving the system design |

**Table 6** **Responsible deployment requirements (continued)**

| OECD Principle | Description | Application to human-AI collaboration deployment |
|---|---|---|
| **Robustness, security and safety** | AI systems must function in a robust, secure and safe way throughout their lifetimes, and potential risks should be continually assessed and managed. | The system must, at all times:<br>▪ Operate safely<br>▪ Protect the user's data<br>In the long-term, the system must:<br>▪ Be maintained, updated and improved<br>▪ Undergo ongoing risk assessments |
| **Accountability** | Organisations and individuals developing, deploying or operating AI systems should be held accountable for their proper functioning in line with the OECD's values-based principles for AI. | The system must have clearly assigned accountability. This includes responsibility for:<br>▪ Human-AI collaboration system deployment<br>▪ Human-AI collaboration system decisions<br>▪ Human-AI collaboration system operation |

These requirements extend the focus of human-AI collaboration beyond immediate task performance. A system may perform effectively in the short term but still create long-term problems if it lacks clear accountability, weakens human rights, introduces unmanaged safety risks, or is not properly maintained.

The design implication is that responsible deployment should be treated as a core requirement from the beginning, rather than as a separate compliance activity after implementation.

# 2.3 The Preconditions to Human-AI Collaboration

### What influences the human-AI collaboration?

The broader socio-technical system of the human-AI collaboration consists of five interconnected domains: the human, AI, task, organisation and society. Each of these domains have preconditions that shape the requirements of the human-AI collaboration system and highlights where design interventions can improve the system.

The purpose of assessing preconditions is not to determine whether human-AI collaboration is simply "ready" or "not ready". Rather, it is to understand where strengths, constraints, and risks exist so that the design can respond to them. For example, limited human experience may increase the need for training and explanation, while high task uncertainty may require greater adaptability and human control.

Desired outcomes and requirements, together with the preconditions that shape them, form the basis for the design decisions discussed next: the human-AI collaboration design principles.

**Figure 5** The five domains of human-AI collaboration preconditions

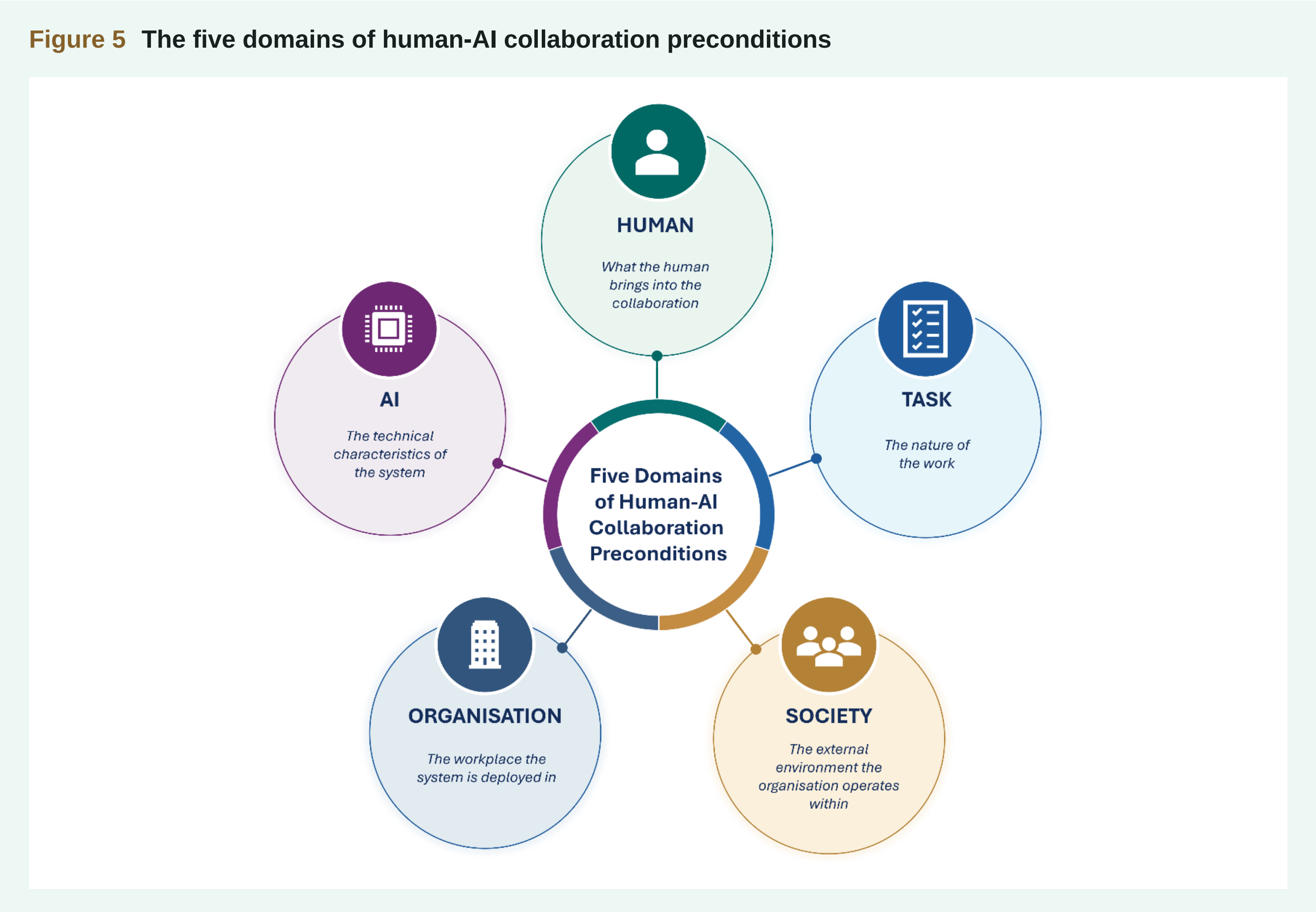

The preconditions of these domains and how to assess these preconditions are listed in Table 7 below.

**Table 7** Preconditions to human-AI collaboration

| Domain | Preconditions | How to assess |
|---|---|---|
| **Human** | ▪ AI & technical literacy [4]<br>▪ Physical & cognitive capability [17]<br>▪ Prior experience [9]<br>▪ Confidence [9]<br>▪ Motivation [17]<br>▪ Personality [17]<br>▪ Communication ability [4]<br>▪ Adaptability [4]<br>▪ Mental capacity & emotional state [8] | ▪ Skills & literacy audit<br>▪ Worker survey |
| **AI** | ▪ Reliability & accuracy [5]<br>▪ Robustness & adaptability [1]<br>▪ Transparency & explainability [3]<br>▪ Usability [9]<br>▪ Safety & privacy protection [1] | ▪ System validation & testing<br>▪ Edge case testing<br>▪ Safety and privacy audit |

**Table 7** **Preconditions to human-AI collaboration (continued)**

| Domain | Preconditions | How to assess |
|---|---|---|
| **Task** | ▪ Complexity [5]<br>▪ Physical & cognitive demands [15]<br>▪ Observability [5]<br>▪ Time pressure [15]<br>▪ Team structure & independence [13]<br>▪ Uncertainty & risk [5] | ▪ Workflow analysis<br>▪ Workload assessment<br>▪ Risk assessment |
| **Organisation** | ▪ Leadership [3]<br>▪ Organisational culture [19]<br>▪ Management support [3]<br>▪ Training provision [4]<br>▪ Change management [11]<br>▪ Resource availability [3]<br>▪ Expected return on investment [3]<br>▪ Incentives [18]<br>▪ Technical support [5]<br>▪ Maintenance [5]<br>▪ Implementation strategy [3] | ▪ Organisation readiness assessment<br>▪ Culture survey<br>▪ Resourcing/budgeting review |
| **Society** | ▪ Legislation [7]<br>▪ Standards [1]<br>▪ Ethics [1]<br>▪ Labour policies [7]<br>▪ Public expectations [18]<br>▪ Professional norms<br>▪ Education systems [7]<br>▪ Economic conditions [7]<br>▪ Public trust [1]<br>▪ Technological maturity[5] | ▪ Regulatory/compliance review<br>▪ Stakeholder consultation<br>▪ Industry standards review |

# 03

# Effective Human-AI Collaboration Design

**How to design effective human-AI collaboration?**

IN THIS CHAPTER

CHAPTER 03

# Effective Human-AI Collaboration Design

Drawing from the conditions that must be satisfied for effective human-AI collaboration (requirements), this section addresses the design decisions that must be made, grouped into six key design principles. These design principles will guide organisations to develop effective human-AI collaboration systems.

## 3.1 The Design Principles

The purpose of the design principles is to turn the human-AI collaboration framework into action. They provide a set of questions that help organisations make deliberate design choices and connect those choices back to the human, teamwork, and responsible deployment requirements identified earlier.

In this way, effective human-AI collaboration is not achieved through one design feature alone, but through a coherent set of decisions that fit the human, AI system, task, organisation, and wider context.

**Table 8 Human-AI collaboration design principles**

| Design Principles | Design Questions |
|---|---|
| **Task & Role Design** | ▪ What role does the AI play? An assistant, autonomous agent?<br>▪ How is the task allocated between the human and AI? Is this allocation fixed or dynamic?<br>▪ How much autonomy and control does the human have over the task and system? |
| **Transparency design** | ▪ To what extent does the human know what the AI observes, decides and does?<br>▪ What level of explanation is provided by AI?<br>▪ What information is provided by default? What information is available on request? |
| **Communication design** | ▪ What channel does the system use to communicate to the human? Voice, screen, gesture?<br>▪ How does the human give feedback to the system, such as commands, confirmations, overrides? |

**Table 8** **Human-AI collaboration design principles (continued)**

| Design Principles | Design Questions |
|---|---|
| **Human-centred design** | ▪ How is trust established early on?<br>▪ What training or onboarding does the human receive?<br>▪ Can the human personalise the system to suit their own preferences?<br>▪ Can the human freely provide suggestions and feedback for improvement and changes? |
| **Responsible design** | ▪ What safety mechanisms are in place?<br>▪ Can the human override, reject or stop the AI at any time?<br>▪ How is human data used and protected?<br>▪ Who is accountable for the deployment, operation and decisions of the system? |
| **Sustainability design** | ▪ What ongoing training and support does the organisation provide?<br>▪ Who is responsible for maintaining and updating the system over time?<br>▪ How does the system fit into existing workflows, tools and processes? |

These principles should not be treated as a fixed checklist or a universal design solution. Their relative importance depends on the requirements and preconditions of each use case. For example, a safety-critical application may require greater attention to human control, transparency, and safeguards, while a highly experienced workforce may need less onboarding but greater flexibility and personalisation.

# 3.2 Applying the framework and design principles

The human-AI collaboration framework is applied through a five-step iterative process that moves from understanding the context to evaluating outcomes.

**Figure 6** **The iterative process to apply the framework and design principles**

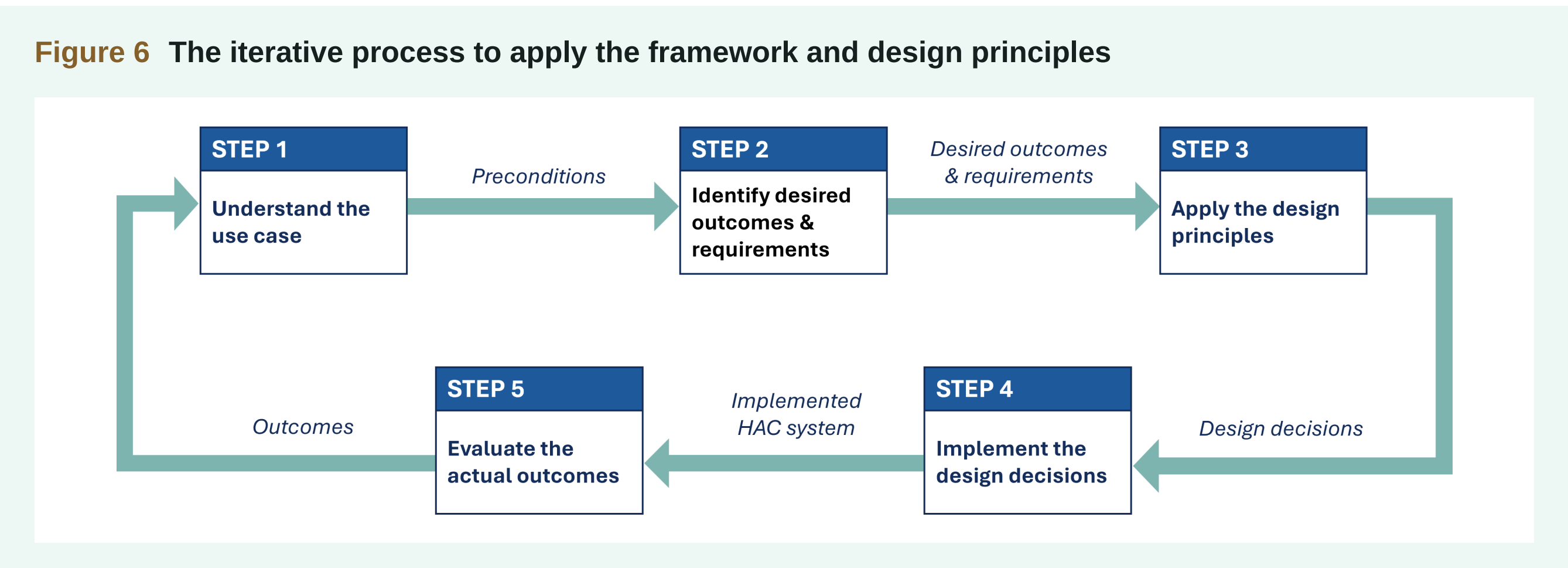

### 01 Understand the use case

Identify the preconditions of the human-AI collaboration system across the human, AI, task, organisation, and societal domains. This establishes the context of the collaboration, including what the human and AI are capable of, what the task requires, what support the organisation can provide, and what external expectations or constraints apply.

### 02 Identify the desired outcomes and requirements

Determine which human needs, teamwork requirements, and responsible deployment requirements are most important for the specific use case. The preconditions identified in Step 1 help determine which requirements should receive the greatest attention.

### 03 Apply the design principles

Use the six human-AI collaboration design principles to make context-specific design decisions. The preconditions indicate which options are feasible and appropriate, while the requirements indicate where the greatest design attention is needed.

### 04 Implement the design decisions

Translate the selected design decisions into the human-AI collaboration system. This may include task allocation, interface design, communication protocols, personalisation options, safety mechanisms, and supporting organisational processes.

### 05 Evaluate the outcomes

Assess the system against the outcomes defined earlier: human outcomes, collaboration and task performance outcomes, and organisational and societal outcomes. The findings can then be used to refine the system and inform the next design cycle.

The five steps should be treated as an iterative process rather than a one-off sequence. As users gain experience, tasks change, AI capabilities evolve, or organisational conditions shift, the preconditions and requirements may also change. The framework can therefore be revisited to support continuous improvement of the human-AI collaboration system.

# 04

# Use Case: Designing and Evaluating a Human-AI Collaborative Assembly System with Cobots

IN THIS CHAPTER

CHAPTER 04

# Use Case: Designing and Evaluating a Human-AI Collaborative Assembly System with Cobots

This use case demonstrates how the human-AI collaboration framework and design principles can be applied to design a human-AI collaboration system and evaluate whether the resulting collaboration is effective.

The chosen collaborative task is a sequential assembly involving the BV704 ball valve manufactured by Oasis Engineering. The worker remained responsible for assembling the valve, while a Franka Research 3 collaborative robot acted as an assistant by supplying the required tools at different stages of the assembly. The system also included an RGB-D camera and action recognition model to detect the current assembly step, together with voice recognition, audio output, and a monitor to support interaction between the worker and robot. Thirty university students were recruited as participants.

The purpose of the use case was not to validate every element of the human-AI collaboration framework. Rather, it was to show how the framework can be translated into design requirements, implemented as specific human-AI collaboration features, and evaluated through a participant study. The study provides sufficient evidence to illustrate human and task-level outcomes, while organisational and societal outcomes are beyond the scope of a single laboratory-based study.

## 4.1 Applying the Human-AI Collaboration Framework to the Use Case

This section follows the five-step process introduced in Section 3.2, applying each step to the collaborative assembly use case, from understanding the use case (Step 1) through to evaluating the actual outcomes (Step 5).

### 4.1.1 Step 1: Understanding the use case

The first step was to identify the preconditions of the collaboration across the human, AI, task, organisation and society domains (Table 9). These establish the context of the use case and inform the desired outcomes, requirements and design decisions in the steps that follow.

**Table 9 Understanding the use case for human-robot collaborative assembly**

| Domain | Use case context |
|---|---|
| **Human** | 30 participants were recruited (mean age 26.7 years, SD 5.8, range 20–44; 23 male, 7 female), predominantly Mechanical or Mechatronics Engineering students. Prior experience varied: most had limited to moderate hand-assembly and AI experience, while the large majority (22 of 30) had no prior experience with collaborative robots |
| **AI** | The AI system consisted of 4 main components:<br>▪ FR3 collaborative robot arm tasked to hand over tools<br>▪ RGB+D camera and recognition model used to detect the current assembly step<br>▪ Voice recognition and speaker setup |
| **Task** | Participants assembled a BV704 ball valve, a mechanical component that controls flow through a pipe. The valve is made up of several parts (body, ball, housing, stem, handle) that must be assembled in the correct sequence, using different tools at different stages. |
| **Organisation** | The use case represents an industrial human-robot collaborative work setting, with minimal training and onboarding provided to the worker. Participants were given briefings, demonstrations and practice attempts to familiarise themselves with the task and AI system prior at the start of each session.<br>The experiments were run at an engineering lab at The University of Auckland. |
| **Society** | The deployment operated within the relevant expectations around worker safety, privacy and ethical use of AI and robotics. All sessions followed The University of Auckland Human Participants Ethics Committee approval. |

### 4.1.2 Step 2: Identifying the desired outcomes and requirements

Given this context, the evaluation covers Human Outcomes and Collaboration and Task Performance Outcomes, both captured through the post-task survey and task data. Organisational and Societal Outcomes are not evaluated, as a single lab-based session cannot speak to productivity, retention or public trust. Because the study is a single session with students rather than industrial workers, short-term, task-level teamwork requirements are the focus. Human needs are also relevant, particularly autonomy and personal well-being (safety and privacy). Responsible deployment requirements are acknowledged but not evaluated.

**Table 10** **The desired outcomes for the human-robot collaborative assembly use case**

| Outcome Category | Desired Outcome in this use case |
| --- | --- |
| **Human Outcomes** | ▪ Workers trust the robot enough to rely on it appropriately (trust & acceptance)<br>▪ Feel in control rather than threatened by it (emotional & psychological)<br>▪ Experience manageable physical and cognitive demand (well-being). |
| **Collaboration and Task Performance Outcomes** | ▪ The valve is assembled efficiently (task performance)<br>▪ The collaboration adapts to the worker's changing preference over repeated use (collaboration quality). |
| **Organisational and Societal Outcomes** | ▪ Not evaluated in this case study. |

**Table 11** **The requirements for the human-robot collaborative assembly use case**

| Requirement Category | Case Study Application |
| --- | --- |
| **Human Needs** | ▪ Psychological: autonomy.<br>▪ Personal and wellbeing: physical safety and privacy |
| **Human-AI Teamwork Requirements** | ▪ Calibrated trust and reliance<br>▪ Appropriate control<br>▪ Adaptability and responsiveness |
| **Responsible Deployment Requirements** | ▪ Treated as a non-negotiable safety and ethics baseline, not evaluated as an outcome. |

## 4.1.3 Step 3: Applying the design principles

By considering the desired outcomes and requirements established in Step 2, the design principles are applied to make design decisions. These design decisions indicate what needs to be implemented in the system so that the human-AI collaboration fulfils the desired outcomes and requirements.

**Table 12 Applying the design principles to the human-robot collaborative assembly use case**

| Design Principle | Design Decision | Implementation |
|---|---|---|
| **Task & Role Design** | ▪ The worker assembles the ball-valve (fixed role).<br>▪ The robot delivers tools (fixed role).<br>▪ Within this, the initiator of the tool delivery is configurable: the worker can request a tool via voice command, or the robot can deliver the tool automatically. | ▪ Worker allocated to assembly<br>▪ Robot allocated to tool handover<br>▪ Configurable handover initiator (human or robot) |
| **Transparency design** | ▪ The robot explains its intended action.<br>▪ Within this, the level of explanation the AI provides is configurable. | ▪ Configurable level of explanations (detailed, brief, none) |
| **Communication design** | ▪ The robot communicates via speech and on-screen text.<br>▪ The worker gives commands via voice command.<br>▪ Recognised commands are displayed on-screen to check for misheard commands. | ▪ Voice input<br>▪ Robot speech output<br>▪ On-screen recognition feedback |
| **Human-centred design** | ▪ Workers receive a briefing, demonstration and practice runs before starting.<br>▪ The system continuously tracks assembly progress and listens for commands.<br>▪ The robot can ask the worker for confirmation before acting. This setting is configurable.<br>▪ Workers can adjust initiator, explanation and confirmation settings. | ▪ Briefing and training<br>▪ Real-time tracking<br>▪ Configurable robot confirmation settings (on or off) |
| **Responsible design** | ▪ The worker can override the robot at any time, regardless of configuration.<br>▪ Researchers supervise every session and can intervene for safety.<br>▪ All safety and ethical requirements were strictly followed and were treated as non-negotiable baseline conditions. | ▪ Continuous human override<br>▪ Supervision |
| **Sustainability design** | ▪ Was not directly tested in this case study | |

### 4.1.4 Step 4: Implement the design decisions

The design decisions in Step 3 were translated into a set of specific configurations and supporting features. Table 13 describes each in detail, including the options presented to participants. Figure 7 shows each implementation in practice.

**Table 13** **Implementing the design decisions for the human-robot collaborative assembly use case**

| Design Decision Implementation | Description |
|---|---|
| **Worker allocated to assembly, robot allocated to tool handover** | ▪ The worker could assemble the assembling valve following their preferred timing. Only instructed to aim for a correct assembly.<br>▪ Tool handover happened over fixed robot trajectories with fixed locations for pick up, confirmation and handover. This was so that the worker could become familiar with the robot faster. Tool handover timing depends on the tool initiator (described below) |
| **Configurable handover initiator** | ▪ Human initiated: the worker requests the tool via voice command.<br>▪ Robot initiated: the robot uses action recognition (an AI technique) to detect the current assembly step and predefined timings to deliver the tool automatically. |
| **Configurable confirmation** | ▪ Confirmation enabled: robot states its intended action and asks for confirmation<br>▪ Confirmation disabled: robot carries out its selected assistance without additional interaction |
| **Configurable explanation** | ▪ Detailed: full reasoning given (e.g., step detected, tool and purpose stated)<br>▪ Brief: action stated only<br>▪ None: no explanation. |
| **Continuous human override** | ▪ Regardless of configuration, the worker could reject, change, or stop the robot's action at any time. |
| **Human voice input and robot speech output** | ▪ The worker gave commands to the system via voice, and a speech-to-text AI model was used to understand what was said.<br>▪ The robot communicated verbally using an AI TTS and speaker. |
| **On-screen recognition feedback** | ▪ Recognised voice commands were displayed on a monitor so the worker could check for misheard input. |
| **Real-time tracking** | ▪ The system continuously monitors the worker’s assembly progress by tracking the unassembled parts on the workbench<br>▪ The system continuously listens for voice commands throughout the task. |
| **Briefing, training and supervision** | ▪ Workers received a task briefing, demonstration and unlimited practice attempts before starting.<br>▪ Two researchers supervised every session and could intervene or override the AI system at any time for safety. |

Figure 7 The implemented design decisions for the human-robot collaborative assembly system

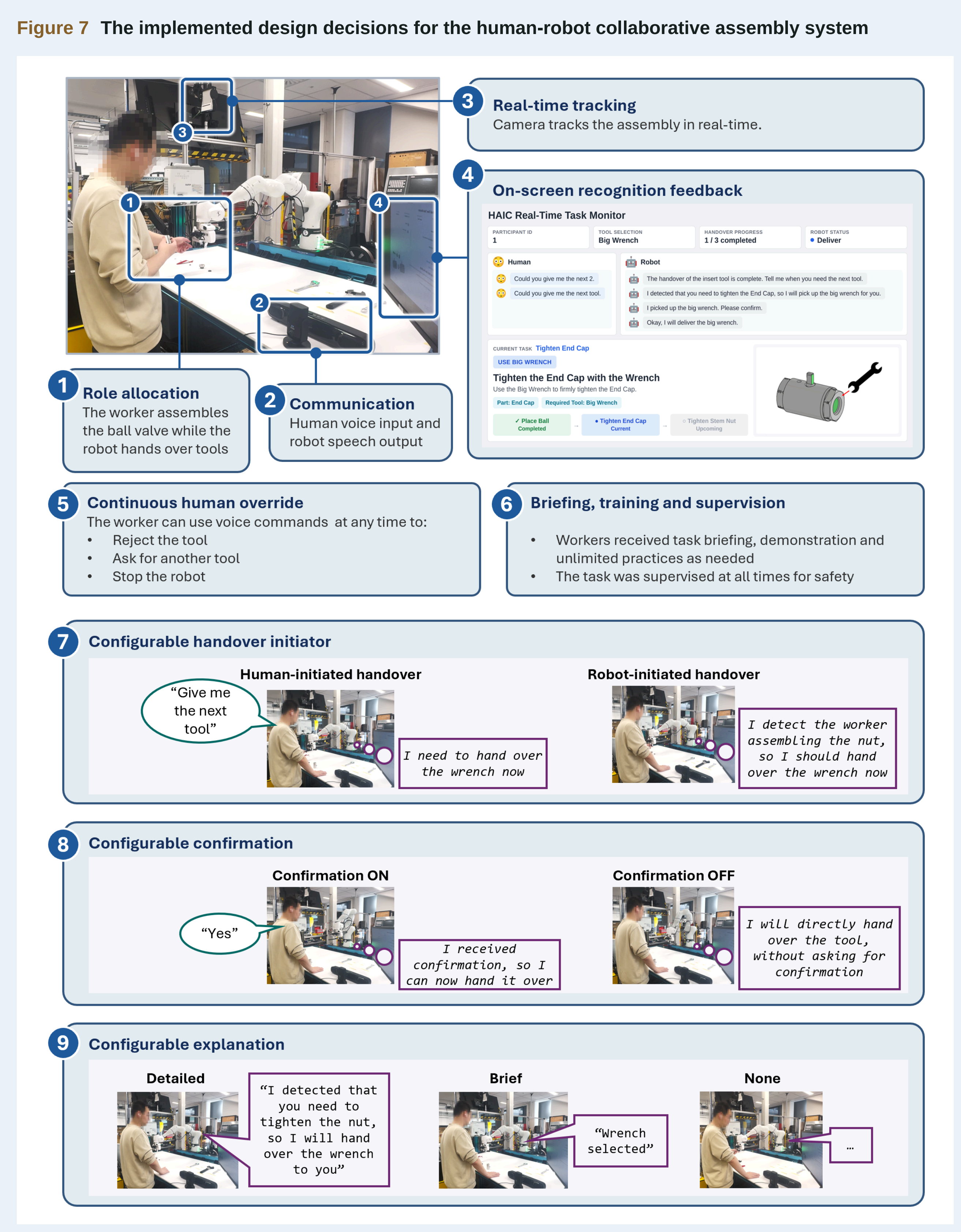

## 4.1.5 Step 5: Evaluate the outcomes

The desired outcomes identified in Step 2 were evaluated using the metrics shown in Table 14. Human outcomes were measured through a survey completed after each assembly run (Appendix). Collaboration and task performance outcomes were measured through system logs, including completion time and the configurations each participant selected.

The metrics were collected through a participant study. Thirty participants each completed six assembly runs. During the first three runs, participants were assigned different combinations of collaboration configurations, exposing them to different forms of initiation, confirmation, and explanation.

During the final three runs, participants were allowed to select their preferred combination of collaboration configurations, enabling the study to examine whether preferred settings changed as participants gained experience with the system.

A controlled error was introduced during Run 5: the robot deliberately handed over the wrong tool. This was used to observe how participants responded to a collaboration failure and whether trust and reliance changed afterward.

**Table 14 Summary of the evaluated outcomes for the human-AI collaborative assembly use case**

| Desired Outcomes (from step 2) | Indicator | Measurement | Actual Outcome |
|---|---|---|---|
| **Human outcomes** | | | |
| **Trust the robot enough to rely on it appropriately** | Trust<br>Reliance | Survey score (1-7 scale) | Trust dropped when the robot made an error but recovered immediately after.<br>Reliance steadily increased throughout the runs, despite drops in Run 5. |
| **Feel in control** | Perceived control | Survey score (1-7 scale) | Perceived control was higher under human-led initiation. However, most participants chose robot-led for efficiency. |
| **Manageable physical and cognitive demand** | Mental demand<br>Physical demand | Survey score (1-7 scale) | Both mental and physical demand decreased across runs (with exception in run 5) |
| **Collaboration and task outcomes** | | | |
| **Product assembled efficiently** | Efficiency | System log: completion time | Completion time decreased across runs (with exception in run 5) |
| **Collaboration adapts to the user's changing preferences** | Configuration changes across runs | System log: configuration settings | 77% of participants changed their configuration at least once, and the system supported each change. |

HUMAN OUTCOMES

## The participants trust the robot enough to rely on it appropriately

Trust stayed fairly stable across the first four runs, dropped sharply in Run 5 when the robot deliberately handed over the wrong tool, and recovered in Run 6. Trust was slightly higher for runs where the participant could choose the configuration (run 4 and 6 averages a score of 5.85) compared to runs where the configuration was assigned for them (run 1-3 average a score of 5.41).

**Figure 8** **Trust and reliance across runs for the human-AI collaborative assembly case study**

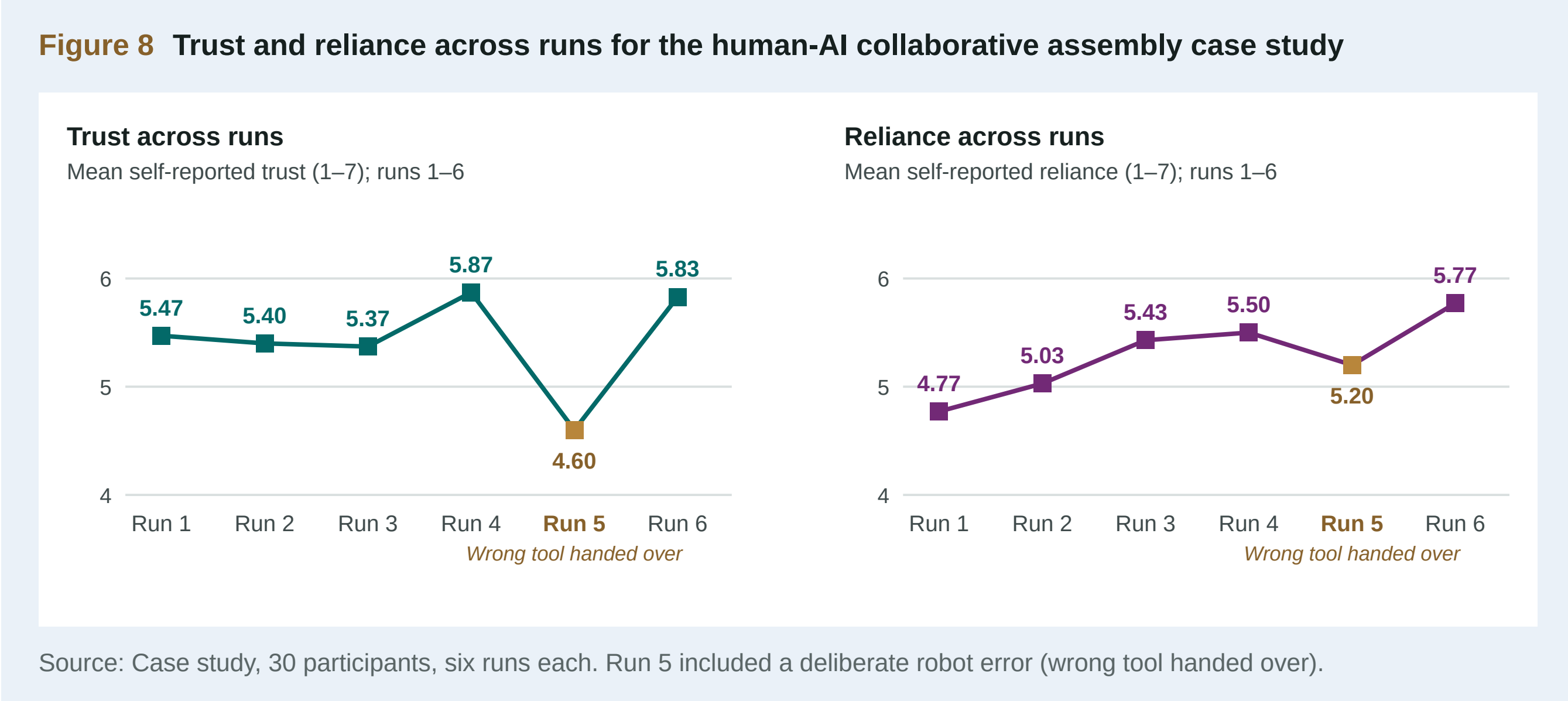

Source: Case study, 30 participants, six runs each. Run 5 included a deliberate robot error (wrong tool handed over).

Trust responded directly to how the robot behaved: one visible error was enough for the trust to drop, and one run of correct behaviour was enough to restore it. This responsiveness is what appropriate trust should look like. Workers adjusted their trust to the robot's actual performance, rather than holding a fixed level of trust regardless of what it did.

Reliance followed a steadier pattern. It rose gradually from 4.77 in Run 1 to 5.77 in Run 6 as participants became familiar with the robot. It dipped only slightly to 5.20 in Run 5.

This steadiness reflects how the task was set up. The robot delivered every tool in every configuration, so participants had to rely on it to complete the assembly regardless of their level of trust. Their reliance was therefore built into the collaboration, not a judgement they made fresh on each run. The rise in reliance over the early runs suggests participants became more comfortable depending on the robot.

## The participants feel in control rather than threatened

Every participant experienced both human-initiated and robot-initiated handover during the assigned runs. The left panel shows the 23 participants who chose a robot-initiated system as their final preference. When comparing the perceived control of a robot vs. human-initiated handover for these participants, they rated their control higher under human-initiated handover (5.58) than under robot-initiated handover (4.37). Hence, despite these participants considering human-initiated handover to give them more control, they chose robot-initiated handover instead.

**Figure 9** **Perceived control of participants for the human-robot collaborative assembly case study**

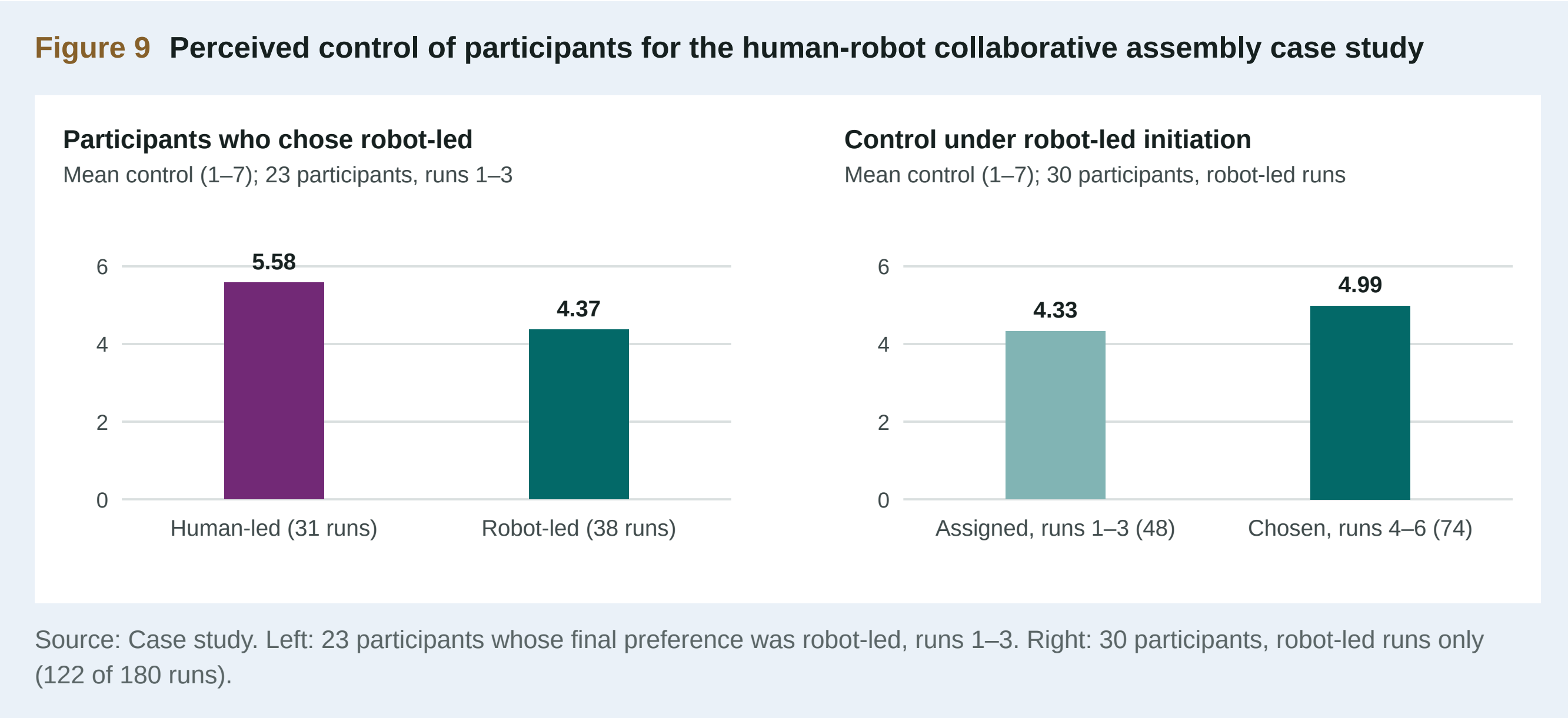

Source: Case study. Left: 23 participants whose final preference was robot-led, runs 1–3. Right: 30 participants, robot-led runs only (122 of 180 runs).

The right panel shows that perceived control under robot-led handovers is not a fixed value. When participants chose the robot-led control themselves in runs 4-6, the perceived control was rated higher (4.99) compared to when the configurations were assigned to them in runs 1-3 (4.33). Being able to choose the configuration appears to add to the participant's sense of control.

Together, these results suggest that the goal is not to maximise control but to ensure the human has enough of it. Once that baseline is met, humans are willing to trade additional sense of control for other benefits such as efficiency.

## Workers experience manageable physical and cognitive demand

Mental demand steadily decreased across the six runs, likely due to being less familiar with the system initially, and slowly gaining familiarity across the runs. There was a temporary rise in Run 5, when participants had to notice and deal with the wrong tool.

**Figure 10** Mental and physical demand of participants for the human-robot collaborative assembly case study

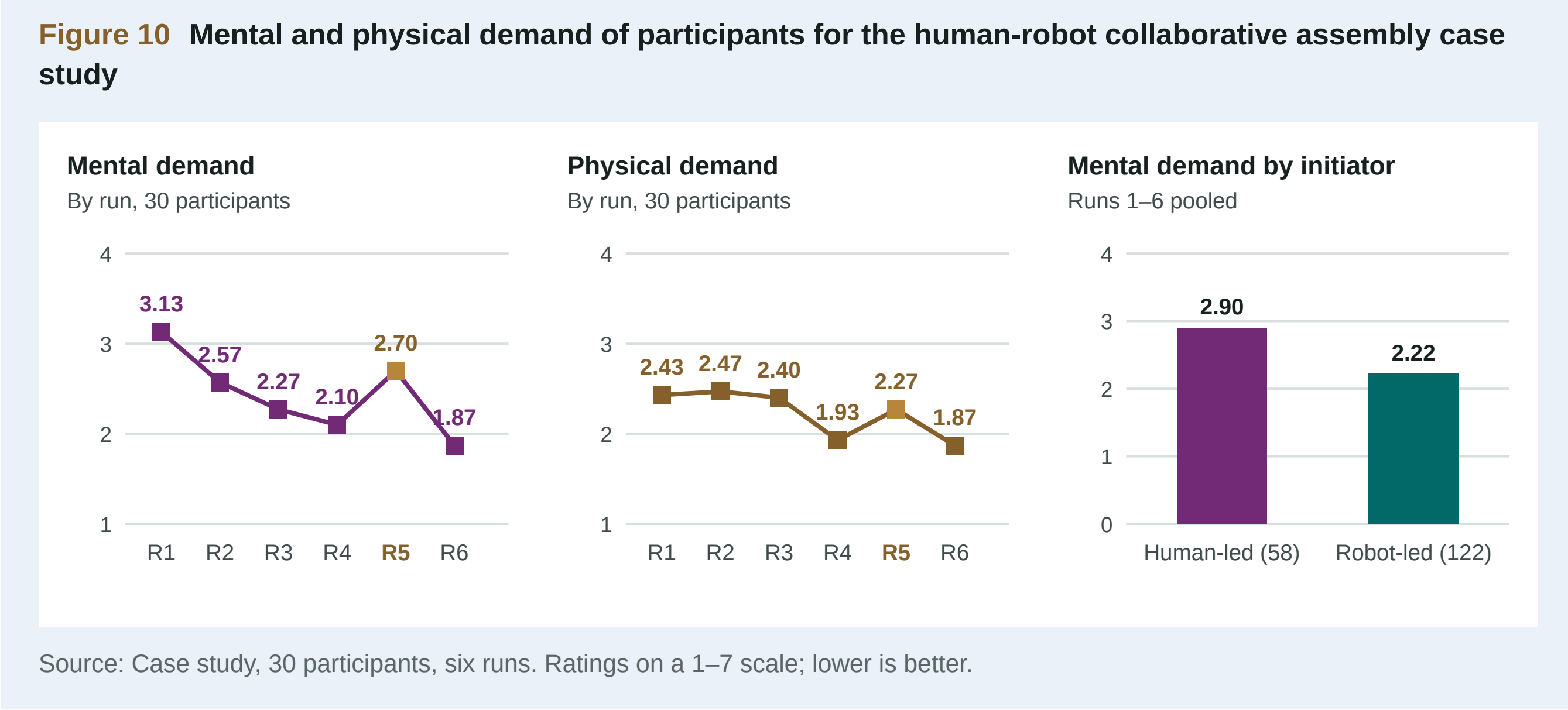

Source: Case study, 30 participants, six runs. Ratings on a 1–7 scale; lower is better.

Physical demand stayed steady at around 2.4 during the assigned runs (Run 1-3) and decreased to around 1.9 once the participants chose their own settings (excluding Run 5).

The configurations affected demand in different ways. Human-led initiation came with higher mental demand than robot-led (2.90 vs 2.22), because the worker had to keep track of which tool was needed and ask for it. Demand responded more to experience than to any single configuration or event, and even the Run 5 error produced only a small, short-lived rise.

## COLLABORATION AND TASK OUTCOMES

### The valve is assembled efficiently

Average completion time fell from 174 seconds in Run 1 to 123 seconds in Run 4, as participants learned the task and moved towards faster configurations. It rose to 157 seconds in Run 5, reflecting the time needed to handle the wrong tool, then reached its lowest point of 118 seconds in Run 6.

Self-chosen runs were faster on average than assigned runs (132 vs 150 seconds, or 120 seconds excluding Run 5). However, this comparison mixes the effect of choosing a configuration with the effect of practice, since the chosen runs always came later.

**Figure 11** Completion time across runs for the human-robot collaborative assembly case study

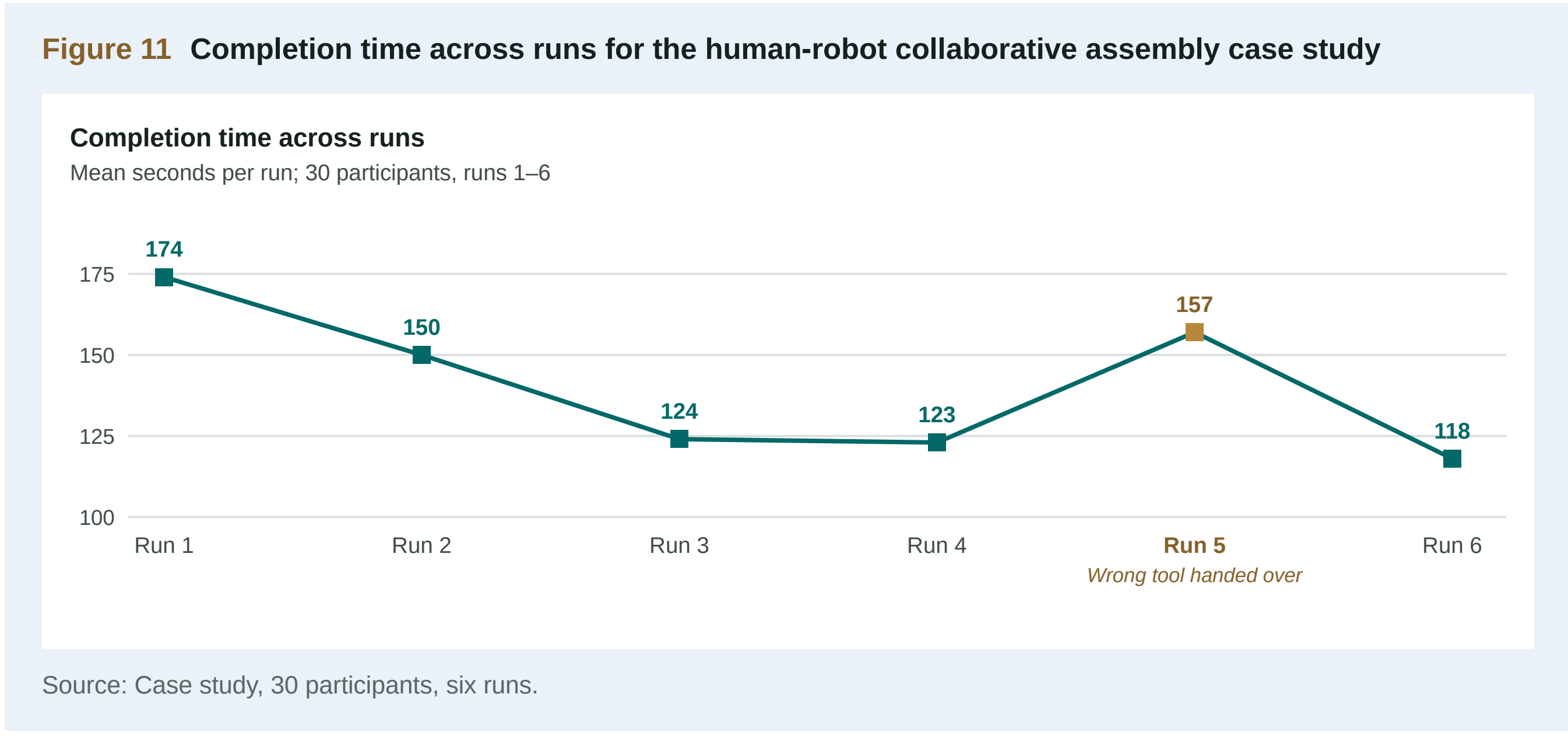

Source: Case study, 30 participants, six runs.

## The collaboration adapts to the worker's changing preferences

By the end of the study, preferences had converged strongly:

**77%**
of participants preferred robot-led initiation

**87%**
preferred confirmation turned off

**29 of 30**
preferred brief or no explanation (all but one)

**Figure 12** Configuration choices of participants for the human-robot collaborative assembly case study

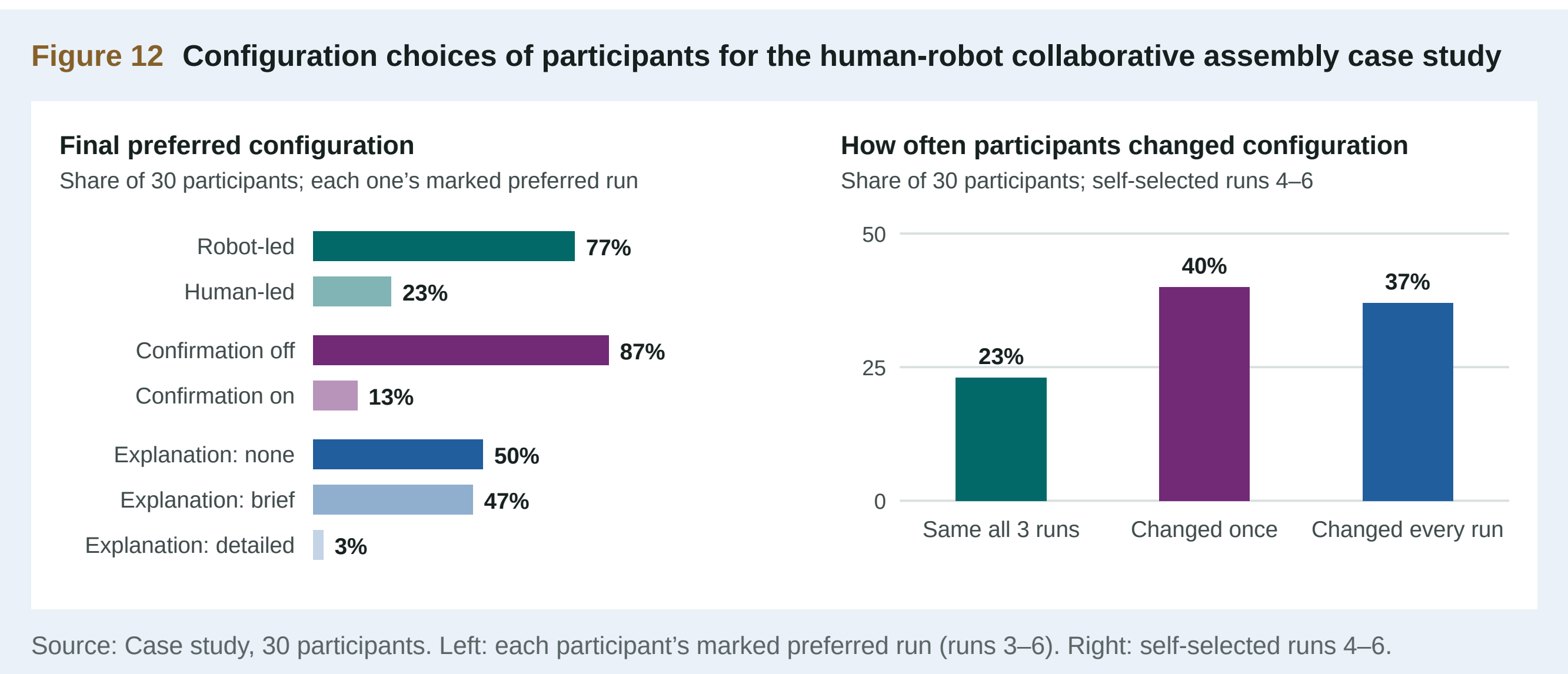

Source: Case study, 30 participants. Left: each participant's marked preferred run (runs 3–6). Right: self-selected runs 4–6.

Although most participants arrived at similar preferences, individuals kept adjusting their settings along the way. Only 23% of participants kept the same configuration across all three self-selected runs. 40% changed once and 37% changed every run. Overall preferences settled quickly, but individual workers continued to refine the exact settings. The system supported this, since participants could change their configuration before each run.

## 4.2 What the Use Case Demonstrates

The use case shows how the human-AI collaboration framework can move from abstract principles to concrete design decisions.

Requirements around autonomy, communication, coordination, transparency, trust, and adaptability were translated into configurable features such as initiation, confirmation, explanation, and human override. These features created different collaboration experiences, which could then be evaluated using both task-performance and human-experience measures.

The results also show why deliberate design matters. The same feature can create both benefits and costs: robot-led initiation improved efficiency but reduced perceived control; confirmation supported oversight but added interaction burden; and detailed explanation increased information while slowing performance.

The case study therefore reinforces a central principle of the human-AI collaboration framework: there is no single universally optimal human-AI collaboration configuration. Effective design requires balancing multiple outcomes and allowing the collaboration to adapt to the worker, task, and context.

> “There is no single universally optimal human-AI collaboration configuration.

# 05

# Implications for Practice

CHAPTER 05

# Implications for Practice

The human-AI collaboration framework is intended to guide practical AI adoption. Its central implication is that organisations should treat human-AI collaboration as a socio-technical design challenge, not simply a technology implementation exercise.

In practice, organisations should:

- Understand the use case first by assessing the human, AI, task, organisational, and societal preconditions that shape the collaboration.
- Prioritise the most important requirements for the context, including worker needs, teamwork quality, safety, transparency, trust, and human control.
- Make design decisions deliberately around task allocation, AI autonomy, communication, explanation, worker control, training, privacy, accountability, and ongoing support.
- Allow the collaboration to adapt over time. The case study showed that workers refined their preferences through repeated interaction and highlighted efficiency, reliability, and personalisation as important areas for improvement.
- Evaluate more than productivity alone. Effective human-AI collaboration should also support positive worker experiences, strong collaboration, and sustainable organisational and societal outcomes.

# Appendix

## Post-Task Survey

Participants completed this survey after each of the six assembly runs (Section 4.1.5). Each item was rated on a 7-point scale.

**Q1** **How mentally demanding was the task?**

1 2 3 4 5 6 7

Very low — Very high

**Q2** **How physically demanding was the task?**

1 2 3 4 5 6 7

Very low — Very high

**Q3** **How rushed did the AI system make you feel?**

1 2 3 4 5 6 7

Not at all — Completely

**Q4** **How much control did you feel you had over the system?**

1 2 3 4 5 6 7

No control — Complete control

**Q5** **How helpful or useful was the AI system?**

1 2 3 4 5 6 7

Not at all useful — Extremely useful

**Q6** **How much did you trust the AI system?**

1 2 3 4 5 6 7

Not at all — Completely

**Q7** **How much did you rely on the AI system?**

1 2 3 4 5 6 7

Not at all — Completely